\documentclass[conference]{IEEEtran}
\usepackage{epsfig}
\usepackage{amsmath}
\usepackage{pdflscape}
\usepackage{graphicx}
\ifCLASSINFOpdf

\else

\fi

\begin{document}

\title{Time-based Mapping of Space Using Visual Motion Invariants}

\author{\IEEEauthorblockN{Juan D. Yepes}
\IEEEauthorblockA{College of Engineering and\\ Computer Science\\
Florida Atlantic University\\
Boca Raton, Florida 33431\\
Email: jyepes@fau.edu}
\and
\IEEEauthorblockN{Daniel Raviv}
\IEEEauthorblockA{College of Engineering and\\ Computer Science\\
	Florida Atlantic University\\
	Boca Raton, Florida 33431\\
	Email: ravivd@fau.edu}}

\maketitle

\begin{abstract}
This paper focuses on visual motion-based invariants that result in a representation of 3-D points in which the stationary environment remains invariant, ensuring shape constancy. This is achieved even as the images undergo constant change due to camera motion. Nonlinear functions of measurable optical flow, which are related to geometric 3D invariants, are utilized to create a novel representation. We refer to the resulting optical flow-based invariants as 'Time-Clearance' and the well-known 'Time-to-Contact' (TTC). Since these invariants remain constant over time, it becomes straightforward to detect moving points that do not adhere to the expected constancy. We present simulations of a camera moving relative to a 3D object, snapshots of its projected images captured by a rectilinearly moving camera, and the object as it appears unchanged in the new domain over time. In addition, Unity-based simulations demonstrate color-coded transformations of a projected 3D scene, illustrating how moving objects can be readily identified. This representation is straightforward, relying on simple optical flow functions. It requires only one camera, and there is no need to determine the magnitude of the camera's velocity vector. Furthermore, the representation is pixel-based, making it suitable for parallel processing.
\end{abstract}

\IEEEpeerreviewmaketitle

\section{Introduction}

% Plan
% We show simulations of the TTC only, Clearance only, individually for stationary environment only
% We show simulations with the combinations
% We will use figures 3.6 (Clearance and TTC visual cues), and figure 3.7 (xyz planes) on the explanation of the method. We relate 3.7 to 3.6.
% Graphically we show the constancy (similar to representations in visual motion 4747)
% We show both TTC-Clearance and XYZ-planes real data KITTI (derotated flow)
% We can talk about dealing with moving objects (some simulations, real data) (say it is beyond the scope of this paper)
% Add to the introduction the green paragraphs (from the word documment sent by Raviv - cgpt transcriptions)

When an observer moves relative to a stationary environment, the projection of 3D objects changes continuously. Interestingly, despite this continuous change, we perceive the world as unchanging and stationary. This prompts a fundamental question: Can we identify certain mathematical characteristics of the image sequence that remain constant through eye movements? In simpler terms, is there an invariant-based representation where transformed projected objects appear unchanged? This enigma of consistent perception amidst ever-changing visual experiences has intrigued researchers over the last eighty years. For instance, Cutting \cite{cutting1986perception} and Gibson \cite{gibson1958visually}, \cite{gibson2014ecological} have both delved into this query, attempting to shed light on its intricacies. According to Gibson, the presence of such invariants, if established, could potentially provide substantial evidence for a groundbreaking theory of perception. Another related part of the puzzle is how we can so easily identify moving objects when the camera itself is in motion. In this sense this paper can be viewed as an extended version of \cite{raviv1992representations}. 

This paper delves into visual motion based mathematical transformations, resulting in a novel time-based representation of 3D objects during the rectilinear movement of a camera. In these representations, a stationary environment remains unchanged or ``frozen," despite continuous alterations in retina images. This is also known as shape constancy \cite{pizlo1994theory}. This representation is constructed using nonlinear functions of optical flow \cite{horn1981determining}\cite{yang2020upgrading}.

Undoubtedly, the fixed geometrical relationships within a stationary environment serve as invariants. Yet, the puzzle is deciphering why we interpret them in this manner given changes in image sequences. What specifically needs to be measured and computed from a sequence of images that gives rise to the perception of stillness?

The representation also helps in solving a highly related problem: It enables the easy identification of moving objects even when the camera is in motion \cite{cutting1995we}\cite{ azim2012detection}.

The new representation is simple, as all measurements within these representations occur in 2D camera coordinates using raw data with no need for 3D reconstruction \cite{ozyecsil2017survey}. Furthermore, this representation is pixel-based, implying that it relies on local computations that can be made in parallel. In this paper, due to limited space, we assume that the camera (observer) undergoes rectilinear motion along its known optical axis.

Three fundamental observations are associated with rectilinear motion relative to a stationary environment: Firstly, the radial distance of a point in 3D space from the camera’s translational trajectory remains constant at any time instant. This implies that all points situated on a specific 3D cylinder, with its axis aligned with the camera’s motion path, maintain an equal distance from the camera’s path at all times. Secondly, the relative depth between any two points in 3D along the translation path remains the same across all time instances. Thirdly, while the camera undergoes rectilinear motion, points on the image plane move radially either away from the Focal Point of Expansion (FOE) or towards the Focal Point of Convergence (FOC) \cite{jain1983direct}.

We first define the coordinate system, then show derivations of two invariants that, when combined, lead to the new representation. These invariants can be computed using non-linear functions of optical flow. This is followed by visually showing a set of images and the resulting constancy in the new domain. We conclude with Unity-based simulation results which visualize the two invariants and the identification of moving objects. 

%\section {Results}
%\subsection{Simulations}
%\subsubsection{Single point simulations}
\section{Method}

\subsection{Coordinate System}
\begin{figure}
	\centering
	{\epsfig{file = 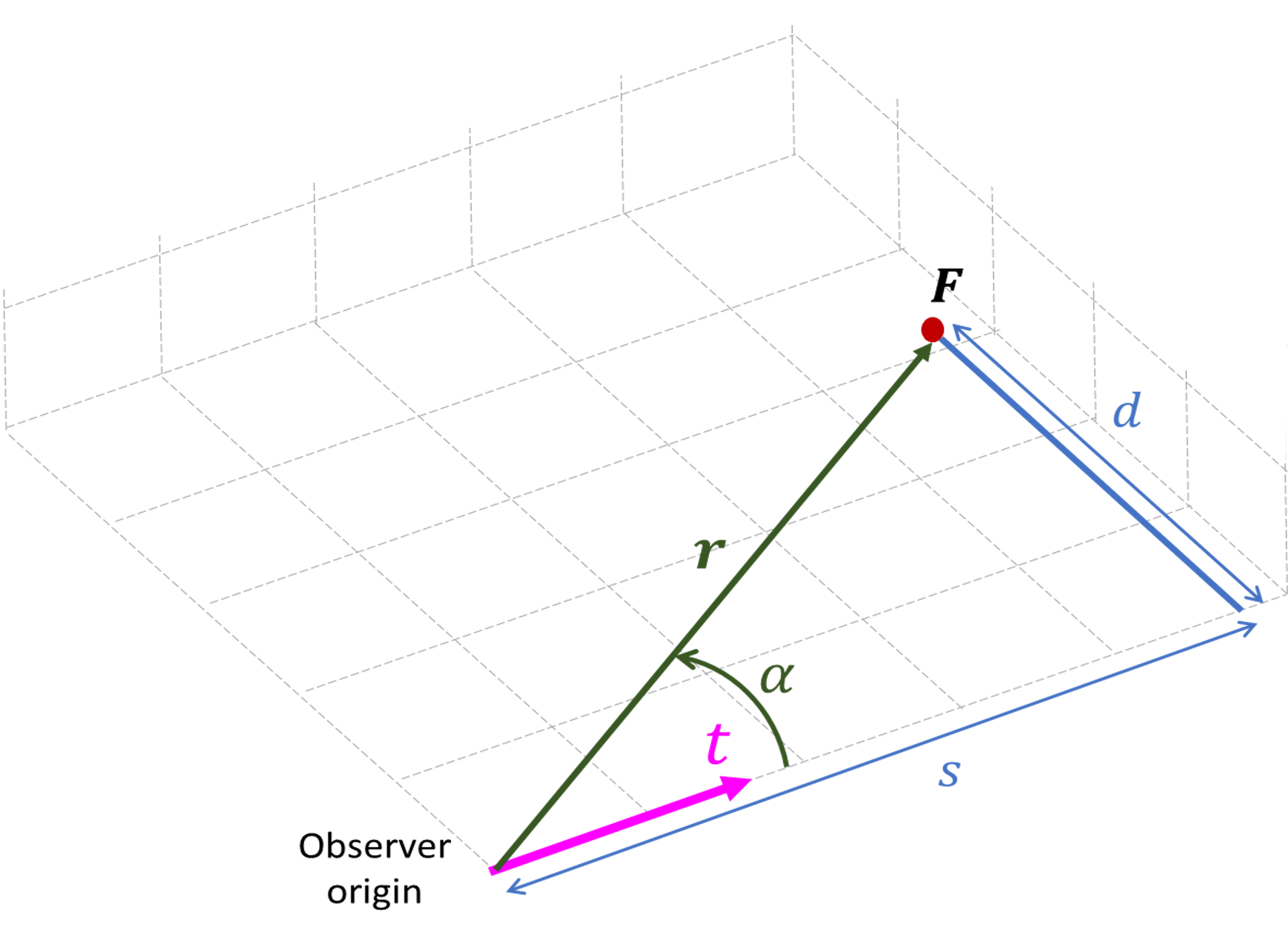, width = 7cm}}
	\caption{A section of the 2D coordinate system.}
	\label{Fig:coordinateSystem}
\end{figure}

\begin{figure}
	\centering
	{\epsfig{file = 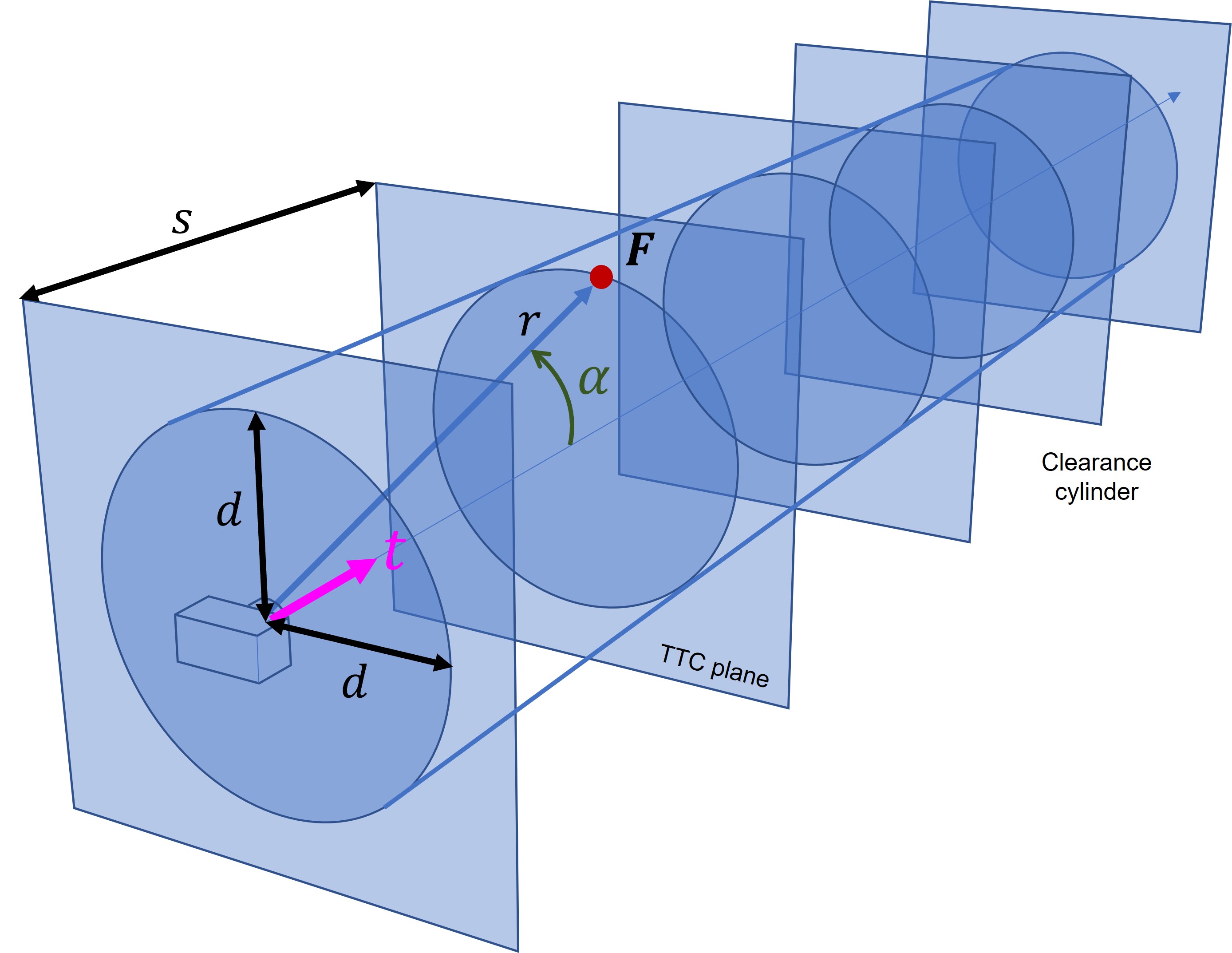, width = 8cm}}
	\caption{3D coordinate system.}
	\label{Fig:cylinder}
\end{figure}

Figure \ref{Fig:coordinateSystem} shows a 2D section of the 3D coordinates relative to the direction of motion vector {\large $\mathbf{t}$}. It can be “rotated” about the optical axis (in our case it is also the direction of motion) depends on the location of the specific point in 3D. 
Refer to Figures \ref{Fig:coordinateSystem} and \ref{Fig:cylinder}. The new representation is defined by three coordinates: Firstly, the physical distance $d$ of a point in 3D from the axis of motion, expressed in terms of optical flow. Secondly, the physical depth $s$ of the point  in 3D  from the camera, also expressed in terms of optical flow. Note $d$ and $s$ define $\alpha$. $\dot{\alpha}$ is the time derivative of $\alpha$. And lastly, the angle of the radial line relative to the horizontal axis of the camera along which the point is moving within the image plane (not shown).
\subsection{Motion Invariants}
Refer to invariants in visual motion in \cite{raviv1993invariants}.\\

\subsubsection{Time-Clearance}
Refer to Figure \ref{Fig:coordinateSystem}. For each point in 3D we have: 
\begin{align}
	d &= r \sin{\alpha}  \notag\\
	r &= \frac{d}{ \sin{\alpha}}.\notag\
\end{align}
From \cite{raviv1993invariants}:
\begin{align}
	\frac{|\mathbf{t}|}{d} &= \frac{\dot{\alpha}}{\sin^{2}(\alpha)} \notag
\end{align}
from which we obtain the Time-Clearance invariant:
\begin{align}
	\text{Time-Clearance} &= \frac{\sin^{2}(\alpha)}{\dot{\alpha}}  \notag	
\end{align}
\subsubsection{Time-to-Contact}
From Figure \ref{Fig:coordinateSystem} we obtain:
\begin{align}
	s &= r \cos{\alpha} \notag\\
	r &= \frac{s}{ \cos{\alpha}}.\notag
\end{align}
From \cite{raviv1993invariants}:
\begin{align}
	\frac{|\mathbf{t}|}{s} &= \frac{\dot{\alpha}}{\cos{\alpha}\sin{\alpha}} \notag
\end{align}
from which we obtain the Time-to-Contact (TTC) invariant \cite{lee2009general}:
\begin{align}
	TTC &= \frac{\sin{2\alpha}}{2\dot{\alpha}}  \notag	
\end{align}

\subsection{Invariants-based Domain}
\begin{figure}
	\centering
	{\epsfig{file = 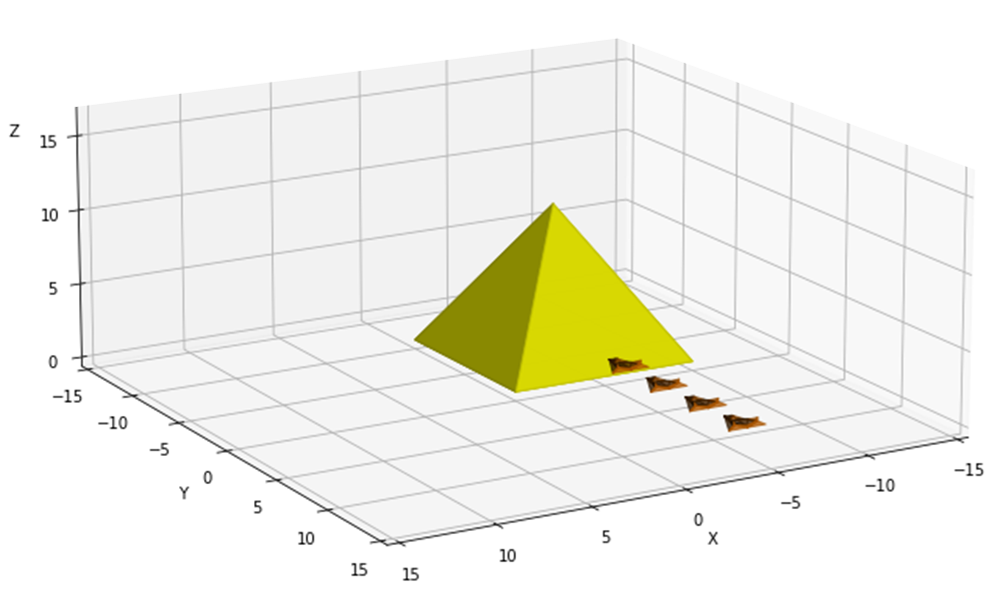, width = 8cm}}
	\caption{World Coordinates with moving camera.}
	\label{Fig:movingCamera}
\end{figure}

Figure \ref{Fig:movingCamera} illustrates a camera, positioned in world coordinates, moving in a straight line at a constant speed towards a 3D object (pyramid).

\begin{figure}
	\centering
	{\epsfig{file = 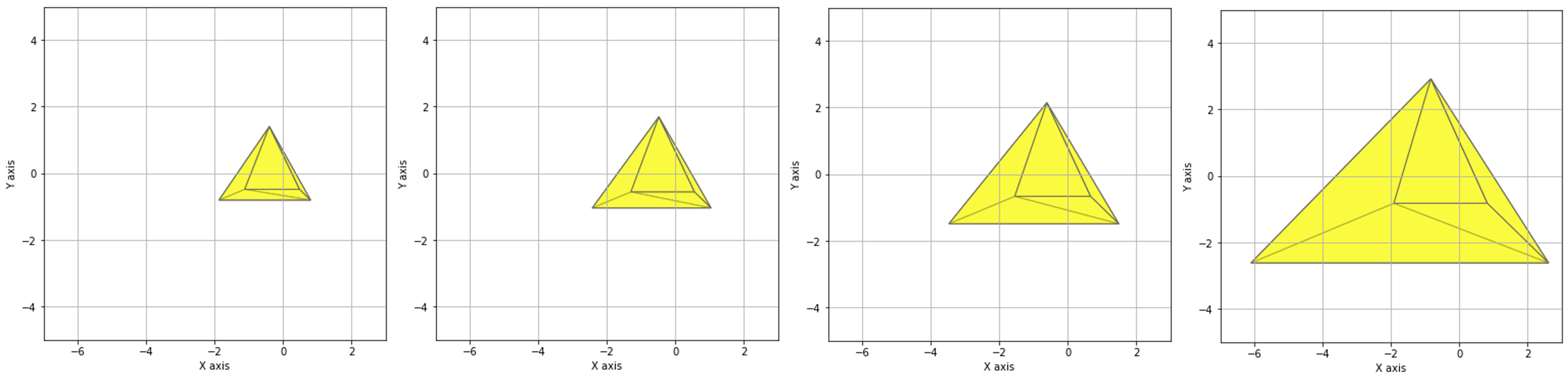, width = 9cm}}
	\caption{2D projections of stationary 3D pyramid as seen by a moving  camera at different time instants}
	\label{Fig:2Dcamera}
\end{figure}

Figure \ref{Fig:2Dcamera} displays the 2D projections captured by the camera at four distinct time instances.

\begin{figure}
	\centering
	{\epsfig{file = 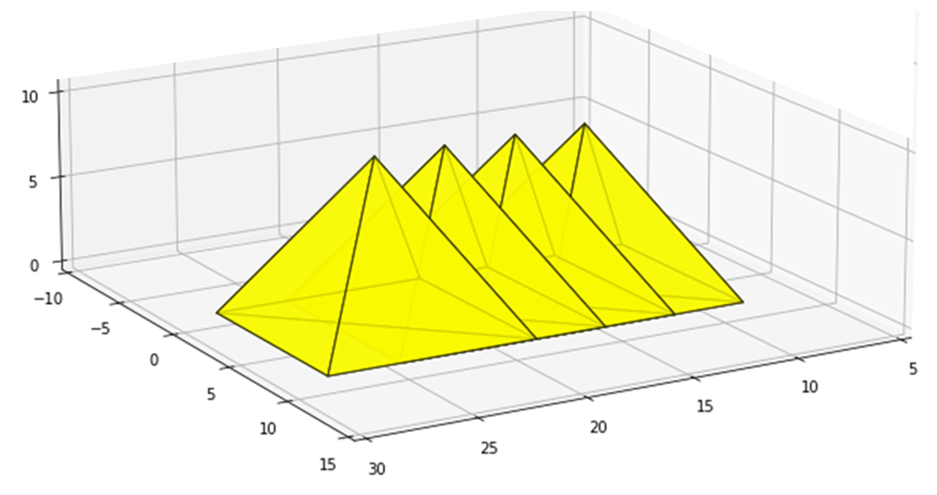, width = 8cm}}
	\caption{Object as obtained in the new representation using Time-Clearance and Time-To-Contact. Note the constancy in the new domain. }
	\label{Fig:newDomain}
\end{figure}

Figure \ref{Fig:newDomain} depicts the object in the new domain at four distinct time instants. Observe the unchanged shape of the object in the new domain, indicating its constancy.

\subsection{Identifying Moving Objects}
In the new representation, moving points in space do not maintain the constant values anticipated in the new domain, and therefore they can be detected there.

\section{Results}
\begin{figure}
	\centering
	{\epsfig{file = 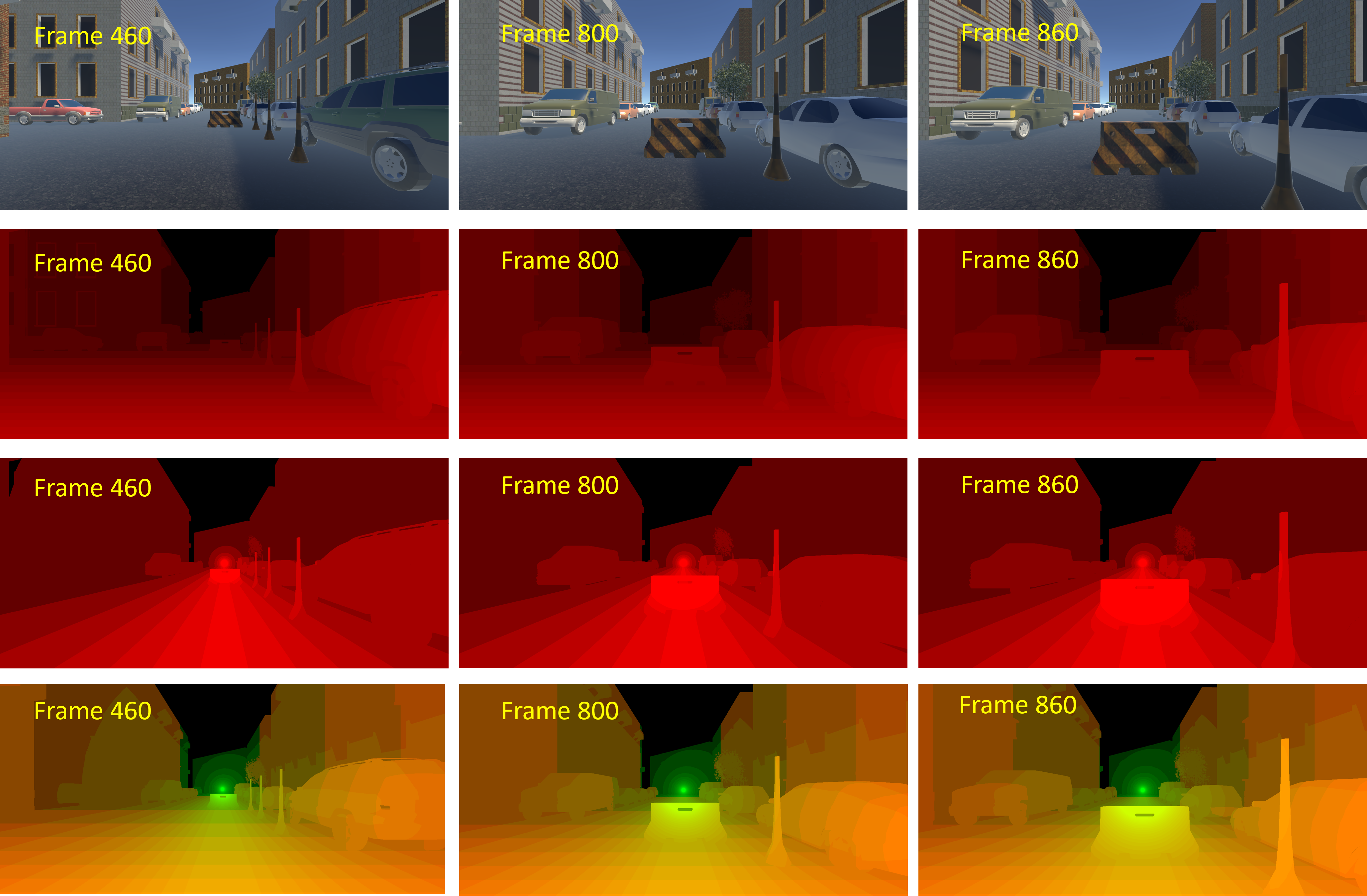, width = 8.5cm}}
	\caption{Color coded visualization of (1/TTC) and (1/Time-Clearance) as obtained from a Unity-simulation sequence for a moving camera in a stationary environment (\textbf{without} moving objects). First row: sample of images at three different time instants. Second row: 1/TTC. Third Row: 1/Time-Clearance. Fourth row: combined visualization of 1/TTC and 1/Time-Clearance.}
	\label{Fig:Unity1}
\end{figure}

\begin{figure}
	\centering
	{\epsfig{file = 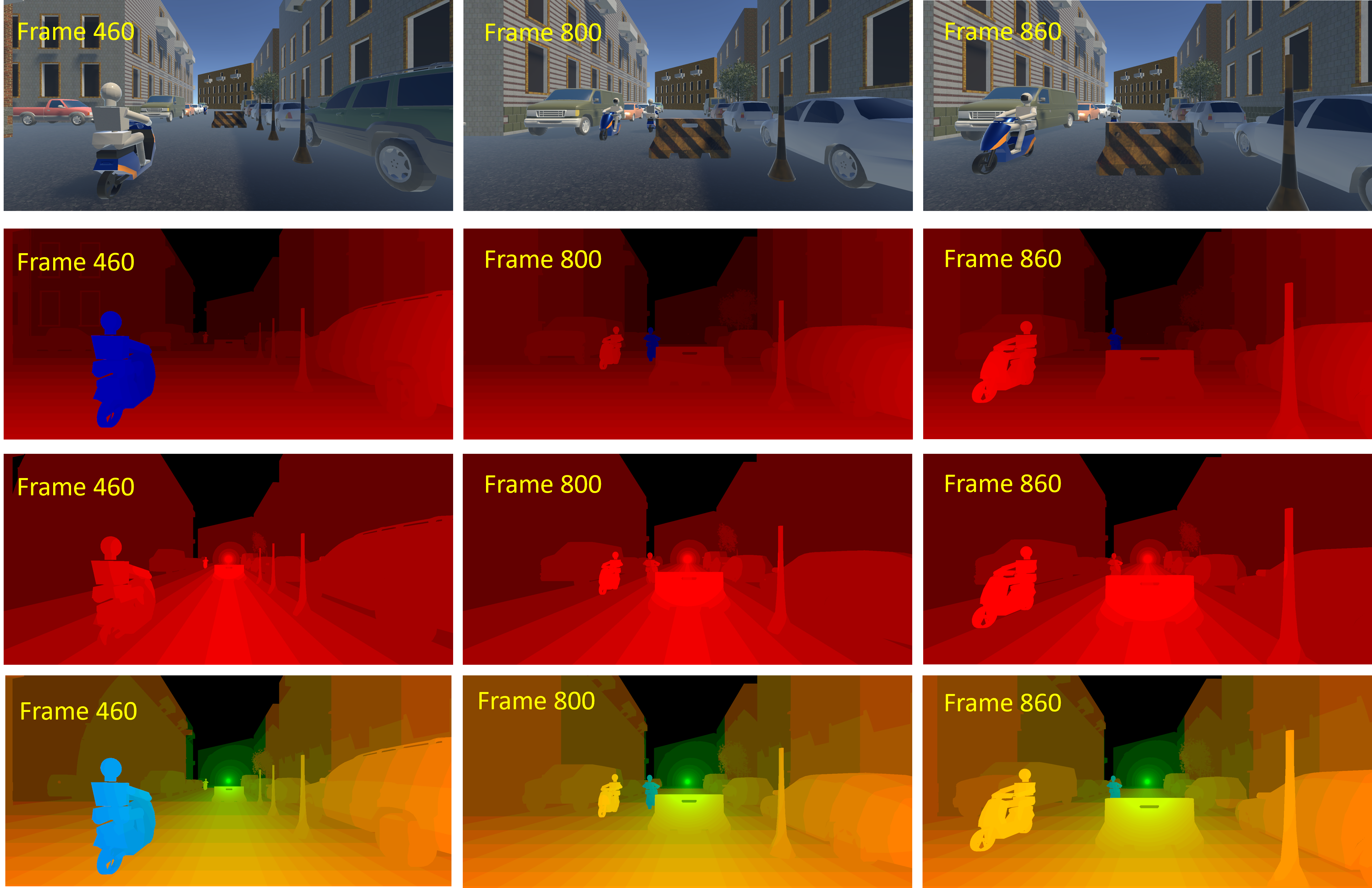, width = 8.5cm}}
	\caption{Color coded visualization of (1/TTC) and (1/Time-Clearance) as obtained from a Unity-simulation sequence for a moving camera in a stationary environment (\textbf{with} moving objects). First row: sample of images at three different time instants. Second row: 1/TTC. Third Row: 1/Time-Clearance. Fourth row: combined visualization of 1/TTC and 1/Time-Clearance.}
	\label{Fig:Unity2}
\end{figure}

This section presents results from two parts of the same simulation. The first set, shown in Figure \ref{Fig:Unity1}, does not include moving objects. The second set, illustrated in Figure \ref{Fig:Unity2}, is identical to the first but with the addition of moving objects that can be clearly identified.

\section{Conclusion}
This paper introduces a new optical flow-based transformation, resulting in a new representation of objects. In this domain, 3D objects seem 'frozen'. This is achieved without needing a 3D reconstruction or prior knowledge about the object. The process is straightforward, suitable for parallel processing, making it ideal for real-time applications. We are also in the process of expanding this method to cater to any 6 degrees of freedom camera motion and any number of moving objects.
\section*{Acknowledgment}
The authors would like to thank M. Levine for his continued support of this project. This work was supported in part at the Technion through a fellowship from the Lady Davis Foundation. We thank M. Herman and J. Albus from NIST. Also thanks to C. Hatcher for suggesting very useful comments.
%\begin{thebibliography}{1}
%\end{thebibliography}
\bibliographystyle{IEEEtran}
\bibliography{IEEEabrv,invariant}
%\end{thebibliography}
\end{document}